\documentclass[runningheads]{llncs}
\usepackage{graphicx}
\usepackage{amsmath,amssymb} % define this before the line numbering.
\usepackage{color}
\usepackage{algorithm}
\usepackage{algpseudocode}
\usepackage{booktabs}
\usepackage{mathtools}
\usepackage{bm}
\usepackage{threeparttable}
\usepackage{multirow}
\usepackage{rotating}
\usepackage{url}

\makeatletter
\newcommand{\printfnsymbol}[1]{%
  \textsuperscript{\@fnsymbol{#1}}%
}
\makeatother

\begin{document}
\pagestyle{headings}
\mainmatter

\def\ACCV20SubNumber{617}  % Insert your submission number here

%===========================================================
\title{Towards Optimal Filter Pruning with Balanced Performance and Pruning Speed} % Replace with your title
\titlerunning{Filter Pruning with Balanced Performance and Pruning Speed}
\author{Dong Li\inst{1,2}\thanks{Contributed equally to this work.} \and
Sitong Chen\inst{2}\printfnsymbol{1} \and
Xudong Liu\inst{2} \and
Yunda Sun\inst{2} \and
Li Zhang\inst{1,2}}
\authorrunning{D. Li et al.}
\institute{Tsinghua University, 100084, Beijing, China \and
Nuctech AI R\&D Center, 100084, Beijing, China\\
\email{\{li.dong, chensitong, liuxudong, sunyunda, zhangli\}@nuctech.com}}

\maketitle

%===========================================================
\begin{abstract}
Filter pruning has drawn more attention since resource constrained platform requires more compact model for deployment. However, current pruning methods suffer either from the inferior performance of one-shot methods, or the expensive time cost of iterative training methods. 
In this paper, we propose a balanced filter pruning method for both performance and pruning speed. Based on the filter importance criteria, our method is able to prune a layer with approximate layer-wise optimal pruning rate at preset loss variation. The network is pruned in the layer-wise way without the time consuming prune-retrain iteration. If a pre-defined pruning rate for the entire network is given, we also introduce a method to find the corresponding loss variation threshold with fast converging speed. Moreover, we propose the layer group pruning and channel selection mechanism for channel alignment in network with short connections. The proposed pruning method is widely applicable to common architectures and does not involve any additional training except the final fine-tuning. Comprehensive experiments show that our method outperforms many state-of-the-art approaches.
\end{abstract}

%===========================================================
\section{Introduction}

Despite the fact that neural network based approaches have achieved significant performance improvement in many computer vision tasks, the deployment of these over-parameterized model often requires high computing power and large memory footprint, which are not available on resource constrained platform such as mobile phone. To tackle this problem, researchers propose different methods for network compression and inference acceleration, including lightweight architecture designing \cite{mobilenet,shufflenet}, network pruning \cite{Optimalbraindamage,Optimalbrainsurgeon,l1norm}, weight quantization \cite{Binaryconnect,Quantized}, matrix factorization \cite{Exploiting}, knowledge distillation \cite{Distilling}, etc.

Among these methods, network pruning has drawn much attention since it is able to reduce the number of model parameters and operations simultaneously. It can be categorized as structure pruning and non-structure pruning. Non-structure pruning sets unimportant weights to zero to achieve high sparsity \cite{hansparse,datafreenos}, while sparse operation requires specialized hardware \cite{hadrware} or software \cite{software} libraries to speed up the inference process, which limits the usage of the pruned network. Structure pruning is also recognized as filter pruning or channel pruning since it is implemented by removing filters in the original network. This coarse-grained filter-level pruning can be treated as modification to the network architecture, so it does not damage the usability of the model. In this paper, we propose a filter pruning method to shrink network size and accelerate its inference at the same time.

The key issue of filter pruning is selecting the unimportant filters to be pruned at a given compression ratio. To solve this combinatorial optimization problem, most methods evaluate the importance of filters then either prune them in a one-shot manner, or iteratively prune-retrain the model. On the one hand, one-shot approaches often prune filters in each layer based on some pre-defined prune rate and particular properties of the trained model \cite{hu,luo,l1norm,cp,geometric,importance}, which are more prone to over-pruning or under-pruning at certain layers. On the other hand, iterative pruning based on greedy criteria increases the time cost and computation burden \cite{geometric,amc,sss}. Also, some of these methods jointly optimize original objective function with compression, thus the loss function becomes more complex and difficult to converge due to the hyper-parameters introduced. Filter pruning is by far an unsolved problem, since the optimal prune rate of each layer is hard to obtain. 

Our approach is a balanced method which is able to approximate the layer-wise optimal pruning rate with limited time and computing resource. Given a trained convolutional network, we observe that removing a convolution kernel in certain layer leads to different changes of loss function and accuracy, while the accuracy drop has a highly positive correlation with the absolute value of loss function change, which we denote it as the loss variation. Based on the criteria of gradient and magnitude of filters, the contribution to loss variation of different filters can be accurately estimated. As these gradients are able to compute by back propagation, we select batches of data to evaluate the importance of each filter per layer by inferences within single epoch. We propose an algorithm to obtain the maximum pruning rate in each layer, constrained by a threshold of loss variation. We use binary search to find the combination of filters which are sorted by importance, so the maximized number of filters can be pruned in one-shot. After all layers are done, the pruned model is fine-tuned only once. To verify the effectiveness of our method, we conduct a series of filter pruning experiments using CIFAR-10 \cite{cifar10} and ImageNet \cite{imagenet} dataset. Our result outperforms the state-of-the-art algorithms with many major network architectures, including VGG \cite{vgg}, GoogLeNet \cite{googlenet}, ResNet \cite{resnet}, DenseNet \cite{densenet}, etc.

In summary, our main contribution is the proposed filter pruning method to approximately obtain layer-wise optimal pruning rate, which is able to prune a layer with maximum pruning rate at given loss variation without the time consuming prune-retrain iteration. For pre-defined pruning rate of entire network, our method is able to converge to the particular pruning rate without additional fine-tuning. We introduce binary search to help the layer-wise pruning and the global pruning rate converging, so that our method balances the performance of pruned network and pruning speed. The proposed method is widely applicable to common architectures of convolutional networks. Comprehensive experiments show that our method is able to achieve higher compression ratio with lower accuracy drop compared with the state-of-the-art approaches.

%===========================================================
\section{Related Work}
Network pruning obtains a more compact model by removing redundant connections from the original network, thereby reducing the number of parameters and operations. Early researches on this topic are mainly addressed by removing weight-level connections for sparse pruning. Since the applicability of sparse network is limited, recent works are more focused on structure pruning methods, which can be further categorized as one-shot filter pruning and iterative filter pruning. 
%----------------------------------------------------------- 
\subsubsection{Sparse Pruning} 
Inspired by neurobiology, the optimal brain damage~\cite{Optimalbraindamage} and the optimal brain surgeon~\cite{Optimalbrainsurgeon} removed unimportant connections according to the analysis of Hessian matrix of the loss function. Han et al.~\cite{hansparse} determined the importance of weights in the network through the weight value, and reduce the redundancy by deleting smaller weights. Srinivas~\cite{datafreenos} proposed a data-free method to remove the redundant parameters of the fully-connected layer. Because of the sparsity of the weight tensor, these unstructured pruned model only accelerate the inference process on specialized platforms.
%----------------------------------------------------------- 
\subsubsection{One-shot Filter Pruning}
refers to all redundant filters in a network are pruned before fine-tuning. Some methods estimated the importance of filters based on the characteristics of the filter itself, including L1 norms~\cite{l1norm}, geometric median~\cite{geometric}, etc. Others evaluated the filter redundancy by analyzing the information of the feature map. Hu et al.~\cite{hu} used the sparsity of the output of each layer to choose the redundant filter. He et al.~\cite{cp} used the least square to reconstruct the error and LASSO regression to remove filters layer by layer. Luo et al.~\cite{luo} pruned filters based on the statistical information of next layer. Yu et al.~\cite{nisp} proposed a method based on importance score propagation, which back-propagates the score of the final response layer to each filter to determine whether the filter is redundant. However, these methods usually depend on a heuristic metric to set the pruning rate of each layer in advance. Although one-shot pruning algorithm is capable to reduce time cost, it is prone to suffer from inferior compression ratio and accuracy.
%----------------------------------------------------------- 
\subsubsection{Iterative Filter Pruning}
selectively prune one or more filters followed by training to recover the model performance in each iteration. Liu et al.~\cite{sliming} performs sparse training on the scale factor of BN, and removes the filter with a smaller scale factor according to the pruning rate corresponding to each layer. Molchanov et al.~\cite{gedient} proposed a criterion based on Taylor expansion to evaluate the importance of filters, then applied with greedy pruning strategy. Reinforcement learning was introduced in AMC~\cite{amc} for pruning, it set rewards by constraining FLOPs, accuracy and specific compression ratios in continuous space. You et al.~\cite{gbn} proposed the Gate Batch Normalization module, and added the FLOPs hyper-parameters to the training objective to compress the model. Huang et al.~\cite{sss} proposed a data-driven method to learn the architecture of the network, introduced a new scaling factor and corresponding sparse regularization, and defined pruning as a joint sparse regularization optimization problem. Lin et al.~\cite{gal} added a mask to each filter and obtained the final model by generative adversarial learning. The dynamic pruning scheme~\cite{gdp} globally pruned unimportant filters and adjusted the network dynamically, with a mechanism to restore the filters that were mis-pruned. Instead of pruning negligible filters, recent work~\cite{centersgd} proposed an optimization objective to generate multiple identical filters then remove them to achieve pruning goals. A major drawback for iterative pruning is the extensive computational burden. Additionally, the pruning strategies based on training iterations often change the optimization function, and even introduce a large number of hyper-parameters, which will make the training more difficult to converge. 

%===========================================================
\section{Our Method}
%-----------------------------------------------------------

\subsection{Filter Importance Evaluation}
\label{section:Importance}
\begin{figure}
	\centering
	\includegraphics[width=0.8\linewidth]{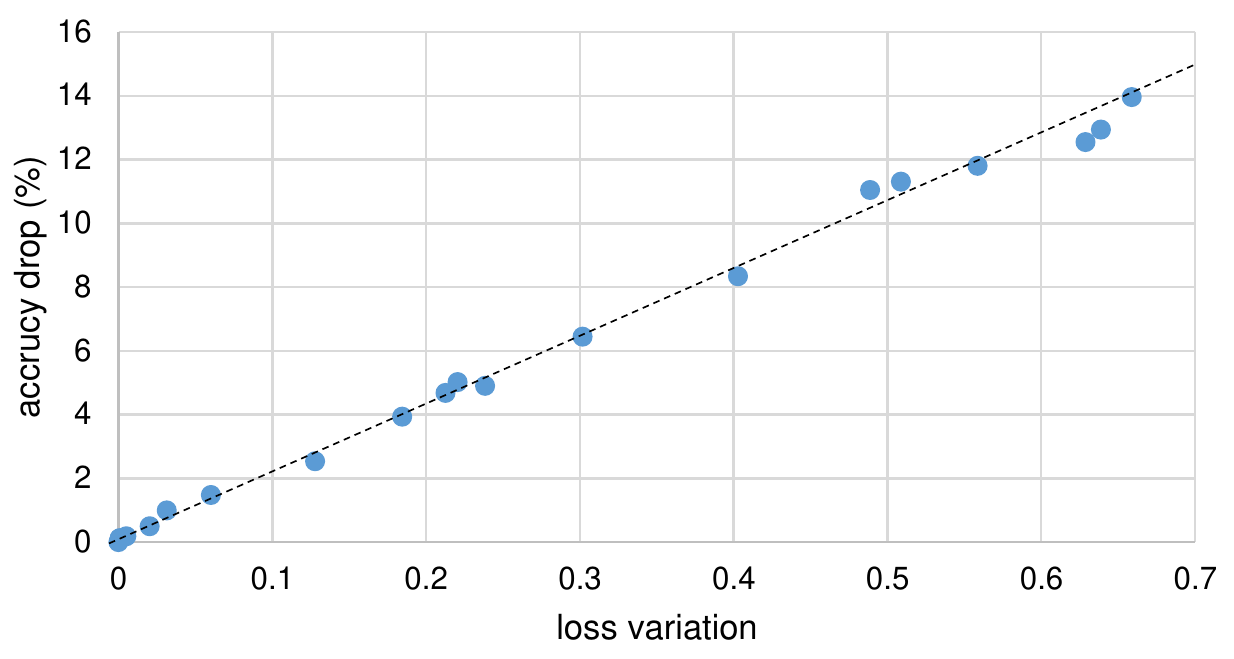}
	\caption{Correlation between loss variation and accuracy drop. Samples are collected from pruning results of VGG16 trained on CIFAR-10.}
	\label{fig:accdrop}
\end{figure}

Given a trained network, we randomly select a group of filters and set their weights to zero, then we use the pruned network to forward all the samples to calculate loss variation and accuracy drop. It is observed that the accuracy drop is almost directly proportional to loss variation. Fig. \ref{fig:accdrop}
has shown the correlation between loss variation and accuracy drop for VGG16 trained on CIFAR-10.

For computer vision application, the majority of trained models are convolutional neural networks. We denote the dataset $\mathcal{D}=\{\mathcal{X},\mathcal{Y}\}$ consists of $N$ samples, $\mathcal{X}=\{\mathbf{x}_0,\mathbf{x}_1,\cdots,\mathbf{x}_N\}, \mathcal{Y}=\{{y}_0,{y}_1,\cdots,{y}_N\}$, where $\mathbf{x}_i$ and $y_i$ are image and label of $i$-th sample, respectively. In a trained network with $L$ layers, the filters can be parameterized as $\mathcal{W}=\{\mathbf{w}^{(1)}_{1},\mathbf{w}^{(2)}_{1},\cdots,\mathbf{w}^{(C_L)}_L\}$, i.e. for $k$-th filter in $l$-th layer, the weights are $\mathbf{w}^{(k)}_{l} \in \mathbb{R}^{C_{l-1}\times p \times p}$ with $l \in [1,2,\cdots,L]$ and $k \in [1,2,\cdots,C_{l}]$, where $C_l$ represents the number of channels in $l$-th layer. We denote the pruned network as $\mathcal{W^{\prime}}$ which sets a subset of filters of $\mathcal{W}$ to zero, i.e. $\mathbf{w}^{(k)}_{l}=\mathbf{0}$ represents the $k$-th filter of $l$-th layer is pruned. Since the accuracy drop is directly related to the loss variation, the filter pruning can be defined as the optimization problem:
\begin{equation}
\begin{aligned}
	&\min _{\mathcal{W}^{\prime}}\left|\mathcal{L}\left(\mathcal{D} ; \mathcal{W}^{\prime}\right)-\mathcal{L}(\mathcal{D} ; \mathcal{W})\right|\\
	&\text { s.t. } \quad\left\|\mathcal{W}^{\prime}\right\|_{0} \leqslant \beta\left\|\mathcal{W}\right\|_{0}
	\label{eq:objective}
\end{aligned}
\end{equation}
where $\mathcal{L}(\cdot)$ is the loss function and $\gamma=1-\beta$ is the specific pruning rate. Solving combinatorial optimization problem (\ref{eq:objective}) is impractical for modern networks, so we evaluate the importance of filters by certain criterion, then use it as the prior knowledge for pruning. The saliency of single filter can be evaluated by calculating the loss variation on the dataset after pruning:
\begin{equation}
\begin{aligned}
	\Delta\mathcal{L}(\mathcal{D} ; \mathcal{W},\mathbf{w}^{(k)}_{l}=\mathbf{0})=\left|\mathcal{L}(\mathcal{D} ; \mathcal{W},\mathbf{w}^{(k)}_{l}=\mathbf{0})-\mathcal{L}(\mathcal{D} ; \mathcal{W})\right|
	\label{eq:gradient}
\end{aligned}
\end{equation}
we have first-order approximation by Taylor expansion:
\begin{equation}
\begin{aligned}
	\left|\mathcal{L}(\mathcal{D} ; \mathcal{W},\mathbf{w}^{(k)}_{l}=\mathbf{0})-\mathcal{L}(\mathcal{D} ; \mathcal{W})\right|	\approx \left|\frac{\partial\mathcal{L}(\mathcal{D} ; \mathcal{W})}{\partial\mathbf{w}^{(k)}_{l}}\mathbf{w}^{(k)}_{l}\right|=\bm{G}^{(k)}_{l}
	\label{eq:taylor}
\end{aligned}
\end{equation}
In this paper, $\bm{G}^{(k)}_{l}$ is used as the criterion to evaluate the importance of $k$-th filter in $l$-th layer. To prune a single filter from a network, the least value of (\ref{eq:taylor}) of all the filters is selected. It is consistent with the intuition that the filter with smaller gradient and magnitude should be pruned first.

%-----------------------------------------------------------
\subsection{Layer-Wise Optimal Pruning Rate Searching}

The value of $\bm{G}^{(k)}_{l}$ is calculated once by forwarding and backwarding all samples from the dataset. In greedy-based methods, the filters are sorted by importance and the unimportant filters are pruned together by their ranks. We propose a layer-wise pruning method to improve the sub-optimal solution caused by cross-layer greedy strategies. 

Suppose there are $ N=M\times P $ samples in the dataset $\mathcal{D}$, where $M$ is the batch size and $P$ is the number of batches. In $l$-th layer, we have $score_{k}$ to represent the importance of $k$-th filter evaluated on the dataset.
\begin{equation}
\begin{aligned}
	score_{k}=\sum_{i=1}^{P}\frac{z_{ik}}{C_{l}} 
	\label{eq:score}
\end{aligned}
\end{equation}
In (\ref{eq:score}), $C_{l}$ is the number of filters in $l$-th layer, $z_{ik}$ is the index of $k$-th filter in ascending order after $i$-th batch running and sorting for $\bm{G}^{(k)}_{l}$. As Sec. \ref{section:Importance} mentioned, the decrease of model accuracy is consistent with the loss variation, so we can use the loss variation as a hyper-parameter for the layer-wise pruning. We treat it as a search problem, aiming to find maximum number of filters to be pruned per layer to achieve high compression ratio within the loss variation range. 
We define the original trained network loss as $\varphi=\mathcal{L}(\mathcal{D} ; \mathcal{W})$ and introduce a parameter $\theta$ that represents the threshold of the loss variation. To accelerate the pruning, we use binary search to avoid the re-evaluation of filter pruning one by one. The algorithm that searches for the optimal pruning rate in one layer is described in Alg. \ref{alg}. The ablation study for Alg. \ref{alg} indecates that the binary search speeds up the pruning process by 5-10$\times$.

\begin{algorithm}[t]
	\caption{Optimal pruning rate searching in $l$-th layer}
	\label{alg}
	\begin{algorithmic}[1]
		\State {\bfseries Input:} original network weights $\mathcal{W}$, $\{score_k\}$ with $k \in [1,2,\cdots,C_l]$, threshold $\theta$, original network loss $\varphi$
		\State {\bfseries Output:} new weights $\mathcal{W^{\prime}}$ with $l$-th layer pruned
		\State $rank \gets C_l/2$
		\State $step \gets rank$
		\State $\{index_k\} \gets \text{sort}(\{score_k\})$ \Comment{in ascending order, $score_{index_1}$ is the smallest}
		\State $prune_{id} \gets \{\}$
		\While {True}
			\If {$step<1$}
				\State {\bfseries break}
			\EndIf
			\State Pruning: $\mathbf{w}^{(i)}_{l} \gets \mathbf{0}$ with $i \in [index_1,index_2,\cdots,index_{rank}]$
			\State Forward all samples to compute the loss $\varphi^{\prime}$ with $l$-th layer pruned
			\State $step \gets step/2$
			\If {$\vert \varphi^{\prime}- \varphi \vert >\theta $}
				\State $rank \gets rank-step$
			\Else
				\State $prune_{id} \gets \{index_k\}$ with $k \in [1,2,\cdots,rank]$
				\State $rank \gets rank+step$
			\EndIf
		\EndWhile
		\State {Pruning: $\mathbf{w}^{(k)}_{l} \gets \mathbf{0}$ for $k \in prune_{id}$}
		\State {\bfseries Return} $\mathcal{W^{\prime}}$
	\end{algorithmic}
\end{algorithm}

\begin{figure}
	\centering
	\includegraphics[width=1.0\linewidth]{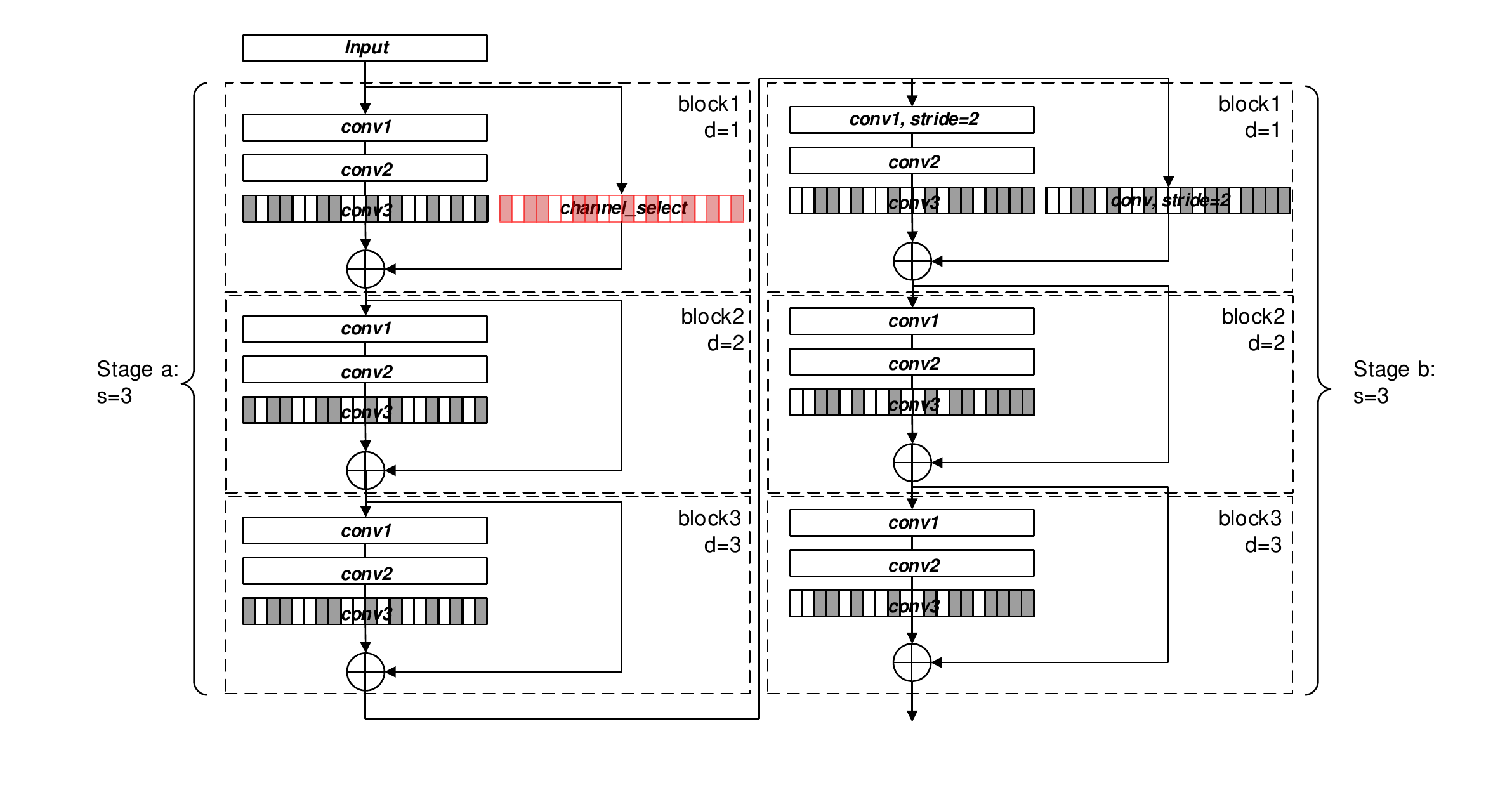}
	\caption {Group pruning method for shortcut connections in ResNet-like networks. For one stage, the last layers of each block are pruned together in a group such that they have the same retained filters. Gray color indicates the corresponding filter is pruned. Since stage a has 3 blocks and the shortcut in each block is an identity connection, a channel\_select layer (in red) is introduced to mask out the pruned channels of input; while stage b has 3 blocks with first shortcut connected by convolutional downsampling, so it can be pruned in a group.}
	\label{fig:network}
\end{figure}

For network architectures with shortcut connections, such as ResNet, the output channels of the last convolutional layers of each block must be the same, as shown in Fig.~\ref{fig:network}. It is because that the shortcut connections require the output channels of these layers are aligned. Therefore, we put the last layers of each block in one stage together as a group for pruning. We define $d$ as the depth of each block in one stage, $s$ denotes the maximum value of $d$. Empirically, we find that the lower-level convolution kernels tend to have weaker representative capability than the high-level ones, so the scoring weight of the lower-level convolution kernels are reduced. The weighted sum for group pruning is conducted by Eq. \ref{score_zh}:
\begin{equation}
\begin{aligned}
	score_{k}=\sum_{i=1}^{K}{\frac{d_i}{s}\cdot score_{k,d_i}}
	\label{score_zh}
\end{aligned}
\end{equation}
where $d_i$ represents depth of $i$-th block in one stage, with $i \in [1,2,\cdots,K]$ and $K$ is the number of blocks. During pruning, the filters with the same index in the group will be pruned together according to $score_k$ ranking. For some stages, the input channels are directly short connected to blocks, hence we introduce a non-parametric layer for channel selection, whose output channel layout is copied from the group pruning result. After the layer-wise optimal pruning rate searching, the layer-wise results are connected for the pruned network, then it is fine-tuned on the training dataset to obtain the final model. 
%-----------------------------------------------------------
\subsection{Pruning with Global Constraints}
In many cases, there are constraints for the whole network, such as global pruning rate $\gamma$ defined in ($\ref{eq:objective}$), performance loss, etc. Here we take pre-defined pruning rate $\gamma$ as the global constraint.
We find that there is a positive correlation between the loss variation and the layer-wise pruning rate, it implies that the loss variation also positively correlated to the global pruning rate. More details of the statistics are listed our supplementary material. We propose a similar binary search method for the loss variation threshold $\theta$, thereby approximate the global pruning constraint $\gamma$. The algorithm is described in Alg. \ref{alg_global}.

Compared to iterative training based pruning methods, our approach only needs fine-tuning the entire network for once. Although pruning with global constraint requires binary search of the loss variation threshold, it only involves multiple times of forwarding, which runs much faster than multiple times of re-training. As the following experiments indicate, our method is able to achieve the balance between performance and pruning speed.

\begin{algorithm}[t]
	\caption{Filter pruning with global pruning rate $\gamma$}
	\label{alg_global}
	\begin{algorithmic}[1]
		\State {\bfseries Input:} original network weights $\mathcal{W}$, $\gamma$, initial value of threshold $\theta_{init}$, pruning rate tolerance $\varepsilon$
		\State {\bfseries Output:} pruned network weights $\mathcal{W^{\prime}}$ with global pruning rate $\gamma^{\prime}$, s.t. $|\gamma^{\prime}-\gamma|\leqslant\varepsilon$
		\State $\theta_{upper} \gets \theta_{init}$
		\State $\theta_{lower} \gets 0$
		\State $\gamma^{\prime} \gets 0$
		\State Forward/backward network $\mathcal{W}$ on $\mathcal{D}$ once, compute original loss $\varphi$ and $\{score_k\}_l$ for $l \in [1,2,\cdots,L]$ and $k \in [1,2,\cdots,C_l]$ \Comment{Eq. \ref{eq:taylor} and \ref{eq:score}.}
		\While {$|\gamma^{\prime}-\gamma|>\varepsilon$}
			\State $\mathcal{W^{\prime}} \gets \mathcal{W}$
			\ForAll {$l$ in $L$ layers}
				\State run pruning Alg. \ref{alg} with ($\mathcal{W}$, $\{score_k\}_l$, $\theta_{upper}$, $\varphi$)
				\State update $\mathcal{W^{\prime}}$ with pruned $l$-th layer
			\EndFor
			\State $\gamma^{\prime} \gets 1-\left\|\mathcal{W}^{\prime}\right\|_{0}/\left\|\mathcal{W}\right\|_{0}$
			\State $step \gets \theta_{upper}-\theta_{lower}$
			\If {$\gamma^{\prime}>\gamma$}
				\State $\theta_{upper} \gets (\theta_{upper}+\theta_{lower})/2$
			\Else
				\State $\theta_{lower} \gets \theta_{upper}$
				\State $\theta_{upper} \gets \theta_{upper}+2\times step$
			\EndIf
		\EndWhile
		\State Finetune the pruned network $\mathcal{L}\left(\mathcal{D} ; \mathcal{W}^{\prime}\right)$ once.
		\State {\bfseries Return} $\mathcal{W^{\prime}}$, $\gamma^{\prime}$, $\theta_{upper}$
	\end{algorithmic}
\end{algorithm}

%===========================================================
\section{Experiments}
We evaluate our algorithm on two commonly used datasets (CIFAR-10 and ImageNet) with popular networks implemented by PyTorch~\cite{pytorch}. The pruning results and performance are compared with several state-of-the-art algorithms in recent years.

%-----------------------------------------------------------
\subsection{Experiments Settings}
\subsubsection{Experiments Setup For CIFAR-10.}
CIFAR-10 is a 10-class image classification dataset containing 50,000 training images and 10,000 test images. In our experiments, we apply VGG16\_BN with a plain structure~\cite{vgg}, GoogLeNet with Inception module~\cite{googlenet}, ResNet-56/110 with residual module~\cite{resnet} and DenseNet-40 with dense connections~\cite{densenet} to verify the effectiveness of our algorithm. For the ResNet-56/110 network on CIFAR-10, the first shortcut of each stage (excluding the first stage) is a downsample and data-filling layer. In order to ensure the channels are aligned, the channel selection layer is added as shown in Fig.~\ref{fig:network}. 
In the fine-tuning phase, we use a NVIDIA Tesla V100 GPU to train the pruned model. We solve the optimization problem by SGD with a Nesterov momentum of 0.9 and weight decay of 1e-4. The network is trained for 400 epochs. The initial learning rate is 0.01, it is decayed by a factor of 10 every 100 epochs. The batch size is 128.
\subsubsection{Experiments setup for ImageNet.}
 ImageNet~\cite{imagenet} is a large image dataset with 1000 classes, containing 1,281,167 training images and 50,000 validation images. In the experiments, we use ResNet-50 to demonstrate our pruning performance on two NVIDIA Tesla V100 GPUs. In the fine-tuning phase, the optimizer parameters are set to be the same as the parameters in CIFAR-10 experiments. The pruned network is fine-tuned for 120 epochs with batch size 256. The initial learning rate is 0.001 and divided by 10 every 30 epochs.
%-----------------------------------------------------------
\subsection{Results on CIFAR-10}
\subsubsection{VGG16\_BN.}
The performance of different compression algorithms are shown in Table.~\ref{tab:cifar10}. PR denotes the pruning rate and FLOPs denotes floating point operations. Ours-0.12 indicates that the threshold of loss variation is 0.12. Compared with L1, SSS, GAL-0.1, and HRank-A, Ours-0.12 has clear advantages for both FLOPs and parameters. Ours-0.12 reduces FLOPs by 70.29\% and deletes 87.72\% of the parameters, while its Top-1 accuracy keeps almost the same as the baseline. For Ours-0.2, although the reductions of FLOPs and parameters are almost the same as those of HRank-B, the Top-1 accuracy is 2.36\% higher than that of HRank-B.
\begin{table}[!t]
	\centering
	\caption{The pruning results on CIFAR-10. L1*, SSS* and ApoZ* are the results in GAL.}
	\centering
	\setlength{\tabcolsep}{2pt}
	{ %
		\renewcommand\tabcolsep{2.5pt} %
		\begin{threeparttable} 
			\begin{tabular*}{\textwidth}{@{}@{\extracolsep{\fill}}clccc@{}}
				\toprule
				\multirow{1}{*}{\textbf{Model}}&{\textbf{Method}} & \textbf{Top-1(\%)} & \textbf{FLOPs(PR)} & \textbf{Parameters(PR)} \\
				\midrule
				{\multirow{8}[0]{*}{{VGG16\_BN}}} &
				Baseline &   {93.96} & {313.73M(0.00\%)} & 14.98M(0.00\%)\\
				& L1*~\cite{l1norm} &  {93.40} & {206.00M(34.34\%)} & 5.40M(63.95\%)\\
				& SSS*~\cite{sss} &  {93.02} & {183.13M(41.63\%)} & 3.93M(73.76\%)\\
				& GAL-0.1~\cite{gal}	& 93.42	 &171.89M(45.21\%) &	2.67M(82.17\%) \\
				& HRank-A~\cite{hrank} &  {93.43} & {145.61M(53.59\%)} & 2.51M(83.24\%)\\
				& \textbf {Ours-0.12} &\textbf{93.95}	 & \textbf{93.22M(70.29\%)} &	\textbf{1.84M(87.72\%)} \\
				& HRank-B~\cite{hrank}	& 91.23 &	73.70M(76.51\%) &	1.78M(88.12\%) \\
				& \textbf {Ours-0.2} &\textbf{93.59} &	\textbf{73.81M(76.47\%)} &	\textbf{1.45M(90.32\%)} \\
				\midrule
				\multirow{9}[0]{*}{{GoogLeNet}} &
				Baseline &	95.05&	1.52B(0.00\%)&	6.15M(0.00\%)\\
				& L1*~\cite{l1norm}&	94.54	&1.02B(32.89\%)&	3.51M(42.93\%)\\
				&Random&	94.54	&0.96B(36.84\%)&	3.58M(41.79\%)\\
				&GAL-0.05~\cite{gal}&	94.56&	0.94B(38.16\%)	&3.12M(49.27\%)\\
				&ApoZ*~\cite{hu} &	92.11	&0.76B(50.00\%)&	2.85M(53.66\%)\\
				&HRank-A~\cite{hrank}	&94.53	&0.69B(54.60\%)&	2.74M(55.45\%)\\
				&\textbf{Ours-0.0045}&\textbf{95.19}	&\textbf{0.57B(62.50\%)}	&\textbf{1.76M(71.38\%)}\\
				&HRank-B~\cite{hrank}	&94.07	&0.45B(70.39\%)	&1.86M(69.76\%)\\
				&\textbf{Ours-0.01}&	\textbf{94.77}&	\textbf{0.40B(73.68\%)}&	\textbf{1.14M(81.46\%)}\\
				\midrule
				\multirow{8}[0]{*}{{DenseNet-40}} &
				Baseline	& 94.81		&282.92M(0.00\%)		&1.04M(0.00\%) \\
				&Liu et al.-40\%~\cite{sliming} &94.81	&190.00M(32.84\%)&	0.66M(36.54\%)\\
				&GAL-0.01~\cite{gal}		&94.61	&	182.92M(35.34\%)		&0.67M(35.58\%)\\
				&HRank-A~\cite{hrank}	&94.24		&167.41M(40.82\%)	&	0.66M(36.54\%)\\
				&Zhao et al.~\cite{zhao} 	&	93.16&	156.00M(44.86\%)	&	0.42M(59.62\%)\\
				&\textbf{Ours-0.02}	&\textbf{94.61}	&\textbf{154.34M(45.45\%)}&\textbf{0.59M(43.27\%)}\\
				&HRank-B~\cite{hrank}		&93.68	&	110.15M(61.07\%)		&0.48M(53.85\%)\\
				&\textbf{Ours-0.04}	&\textbf{93.49}	&\textbf{95.69M(66.18\%)}&\textbf{0.37M(64.42\%)}\\
				\midrule
				\multirow{9}[0]{*}{{ResNet-56}} &
				Baseline &	93.26 &	125.49M(0.00\%)&	0.85M(0.00\%) \\
				&L1*~\cite{l1norm}	& 93.06 &	90.90M(27.56\%)&	0.73M(14.12\%)\\
				&NISP~\cite{nisp}&	93.01&	81.00M(35.45\%)&	0.49M(42.35\%)\\
				&HRank-A~\cite{hrank}&	93.17&	62.72M(50.02\%)&	0.49M(42.35\%)\\
				&He et al.~\cite{cp}&	90.80&	62.00M(50.59\%)&	-\\
				&\textbf	{Ours-0.019}&	\textbf{93.64}&	\textbf{59.84M(52.31\%)}&\textbf{0.52M(38.82\%)}\\
				&GAL-0.8~\cite{gal}&	91.58&	49.99M(60.16\%)&	0.29M(65.88\%)\\
				&HRank-B~\cite{hrank}&	90.72&	32.52M(74.08\%)&	0.27M(68.24\%)\\
				&\textbf{Ours-0.055}&	\textbf{91.54}&	\textbf{25.72M(79.50\%)}&\textbf{0.25M(70.59\%)}\\
				\midrule
				\multirow{5}[0]{*}{{ResNet-110}} &
				Baseline&	93.50&	252.89M(0.00\%)&	1.72M(0.00\%)\\
				&L1*~\cite{l1norm}&	93.30 &	155.00M(38.71\%)&	1.16M(32.56\%)\\
				&GAL-0.5~\cite{gal}&	92.55&	130.20M(48.52\%)&	0.95M(44.77\%)\\
				&HRank~\cite{hrank}&	93.36&	105.70M(58.20\%)&	0.70M(59.30\%)\\
				&\textbf{Ours-0.007}&\textbf{93.73}&	\textbf{98.04M(61.23\%)}&\textbf{0.89M(48.26\%)}\\
				\bottomrule
			\end{tabular*}%
		\end{threeparttable}
	}
	\label{tab:cifar10}%
\end{table}%

\subsubsection{ResNet56/110.} The results for ResNet56/110 are shown in Table.~\ref{tab:cifar10}.
Firstly, we look into the result of ResNet56. 
Compared with L1, Ours-0.019 obtains more FLOPs and parameters reductions with higher Top-1 accuracy, and even its accuracy is 0.38\% higher than the baseline.
Although Ours-0.019 has a slightly lower pruning rate than NISP and HRank-A, it achieves larger reductions in FLOPs 
(52.31\% vs. 35.45\% by NISP and 52.31\% vs. 50.02\% by HRank-A)
and better Top-1 accuracy (93.64\% vs. 93.01\% by NISP and 93.47\% vs. 93.17\% by HRank-A).
Therefore, it can be verified that our method is able to greatly reduce the amount of calculation and memory footprint while achieves better model performance. 
From the result, Ours-0.055 can obtain a network with higher compression ratio, its FLOPs is reduced by 79.50\%, parameters are pruned by 70.59\%, at the cost of Top-1 accuracy drop by only 1.72\%. Compared with GAL-0.8, Ours-0.055 achieves much higher pruning rate of FLOPs and parameters though their accuracies are almost the same.
Meanwhile, it outperforms HRank-B in all three aspects.

Next, we analyze the result of ResNet110.
Ours-0.007 leads to an improvement in Top-1 accuracy over the baseline model
(93.73\% vs. 93.50\%) with 61.23\% FLOPs and 48.26\% parameters reductions.
Its performance is significantly better than L1 and GAL-0.5. 
Compared with HRank, Ours-0.007 achieves higher reduection rate of FLOPs (61.23\% by Ours-0.007 vs. 58.20\% by HRank) and better accuracy (93.73\% by Ours-0.007 vs. 93.36\% by HRank), although the parameter pruning rate is lower.

\subsubsection{DenseNet-40.}Table.~\ref{tab:cifar10} summarizes the result of DenseNet-40. Although Liu et al.~\cite{sliming} retains the same accuracy as the baseline, the compression ratio is relatively low, reducing FLOPs by only 32.84\%.
For Ours-0.02, 45.45\% of FLOPs and 43.27\% of the parameters are reduced, and the decrease of accuracy is only 0.20\%. Ours-0.02 achieves a better performance compared with HRank-A and Zhao et al~\cite{zhao}.
Compared with GAL-0.01, Ours-0.02 has a big gain on both FLOPs and parameters pruning rate with the same Top-1 accuracy.
For Ours-0.04, the Top-1 accuracy is 0.19\% lower than that of HRank-B, but our method obtains more reductions of FLOPs and parameters.

\subsubsection{GoogleNet.}The results of GoogleNet are shown in Table.~\ref{tab:cifar10}. 
Ours-0.0045 obtains 95.19\% Top-1 accuracy, which is even 0.14\% higher than the baseline, and 64.47\% of FLOPs and 73.17\% of parameters are removed. It outperforms L1, Random, GAL-0.05, APoZ and HRank-A.
Furthermore, we set the threshold to 0.01 to increase the pruning rate of the network.
Ours-0.01 achieves a better performance than HRank-B (94.77\% acc vs. 94.07\% by HRank-B, 73.68\% reduction of FLOPs vs. 70.39\% by HRank-B, 81.46\% reduction of parameters vs. 69.76\% by HRank-B). 

%-----------------------------------------------------------
\subsection{Results on ImageNet}
Experiments are also conducted on the ImageNet dataset using ResNet50, and the results are shown in Table.~\ref{tab:imagenet}. As indicated by the results, our method has achieved a significant gain on both performance and compression ratios compared with several state-of-the-art methods. 
Specifically, we set the thresholds to 0.05, 0.09, 0.2 and 0.35 respectively to obtain different pruning rates.
Ours-0.05 outperforms GAL-0.5, SSS-26 and HRank-A. It removes 43.03\% FLOPs from baseline, while still yields 75.79\% Top-1 accuracy and 92.82\% Top-5 accuracy, improves the result of SSS-32~\cite{sss} and He at al.~\cite{cp} by a large margin. In addition, Ours-0.09 achieves 75.04\% Top-1 accuracy and 92.29\% Top-5 accuracy with 53.30\% and 40.16\% reductions of FLOPs and parameters, respectively. 
Moreover, compared to GDP-0.6, GAL-0.5-joint, GAL-1, GDP-0.5 and HRank-B, Ours-0.2 has apparent advantages in all aspects, including Top-1/Top-5 accuracy as well as FLOPs and parameters reductions.
For Ours-0.35, 75.80\% FLOPs and 68.55\% parameters are removed, its 70.58\% Top-1 accuracy and 90.00\% Top-5 accuracy are significantly better than those of GAL-1-joint and ThiNet-50. Compared with HRank-C, Ours-0.35 achieves higher Top-1 and Top-5 accuracy with the similar FLOPs and parameters reductions.
Therefore, the ImageNet experiments indicate that our method also works well on large and complex datasets.

\begin{table}[!t]
	\centering
	\caption{Pruning results of Resnet-50 on ImageNet.}
	\centering
	\setlength{\tabcolsep}{2pt}
	{ %
		\renewcommand\tabcolsep{2.5pt} %
		\begin{threeparttable} 
			\begin{tabular*}{\textwidth}{@{}@{\extracolsep{\fill}}clcccc@{}}
				\toprule
				\quad&\textbf{Method}& \textbf{Top-1(\%)} & \textbf{Top-5(\%)} & \textbf{FLOPs(PR)} & \textbf{Parameters(PR)} \\
				\midrule
				\quad&Baseline&	76.15&	92.87&	4.09B(0.00\%)&	25.50M(0.00\%)\\
				\quad&SSS-32~\cite{sss}	&74.18&	91.91&	2.82B(31.05\%)&	18.60M(27.06\%)\\
				\quad&He et al.~\cite{cp}&	72.30&	90.80&	2.73B(33.25\%)&	-\\
				\quad&\textbf{Ours-0.05}&	\textbf{75.79}&	\textbf{92.82}&	\textbf{2.33B(43.03\%)}&	\textbf{17.93M(29.69\%)}\\
				%\midrule
				\quad&GAL-0.5~\cite{gal}&	71.95&	90.94&	2.33B(43.03\%)&	21.20M(16.86\%))\\
				\quad&SSS-26~\cite{sss}&	71.82&	90.79&	2.33B(43.03\%)&	15.60M(38.82\%)\\
				\quad&HRank-A~\cite{hrank}	&74.98&	92.33&	2.30B(43.76\%)&	16.15M(36.67\%)\\
				\quad&\textbf{Ours-0.09}&	\textbf{75.04}&	\textbf{92.29}&	\textbf{1.91B(53.30\%)}&	\textbf{15.26M(40.16\%)}\\
				%\midrule
				\quad&GDP-0.6~\cite{gdp}	&71.19&	90.71&	1.88B(54.03\%)&	-\\
				\quad&GAL-0.5-joint~\cite{gal}&	71.80&	90.82&	1.84B(55.01\%)&	19.31M(24.27\%)\\
				\quad&GAL-1~\cite{gal}&	69.88&	89.75&	1.58B(61.37\%)&	14.67M(42.47\%)\\
				\quad&GDP-0.5~\cite{gdp}&	69.58&	90.14&	1.57B(61.61\%)&	-\\
				\quad&HRank-B~\cite{hrank}	&71.98&	91.01&	1.55B(62.10\%)&	13.77M(46.01\%)\\
				\quad&\textbf{Ours-0.2}&	\textbf{73.06}&	\textbf{91.30}&	\textbf{1.31B(67.97\%)}&	\textbf{10.84M(57.49\%)}\\
				%\midrule
				\quad&GAL-1-joint~\cite{gal}&	69.31&	89.12&	1.11B(72.86\%)&	10.21M(59.96\%)\\
				\quad&ThiNet-50~\cite{luo}&	68.42&	88.30&	1.10B(73.11\%)&	8.66M(66.04\%)\\
				\quad&HRank-C~\cite{hrank}&	69.10&	89.58&	0.98B(76.04\%)&	8.27M(67.59\%)\\
				\quad&\textbf{Ours-0.35}&	\textbf{70.58}&	\textbf{90.00}&	\textbf{0.99B(75.80\%)}&	\textbf{8.02M(68.55\%)}\\
				\bottomrule
			\end{tabular*}%
		\end{threeparttable}
	}
	\label{tab:imagenet}%
\end{table}%
\begin{table}[!t]
	\centering
	\caption{Pruning results of SSD on VOC0712.}
	\centering
	\setlength{\tabcolsep}{2pt}
	{ %
		\renewcommand\tabcolsep{2.5pt} %
		\begin{threeparttable} 
			\begin{tabular*}{\textwidth}{@{}@{\extracolsep{\fill}}clccc@{}}
				\toprule
				\quad&\textbf{Method}& \textbf{mAP(\%)} & \textbf{FLOPs(PR)} & \textbf{Parameters(PR)} \\
				\midrule
				\quad&Baseline&  77.68&  31.40B(0.00\%)&  26.29M(0.00\%)\\
				\quad&Ours-0.05&  77.78&  21.69B(30.92\%)&  17.14M(34.80\%)\\
				\quad&Ours-0.1&  77.14&  18.27B(41.82\%)&  14.17M(46.10\%)\\
				\quad&Ours-0.3&  76.45&  12.22B(61.08\%)&  9.29M(64.66\%)\\
				\quad&Ours-0.5&  75.83&  9.69B(30.92\%)&  6.87M(16.67\%)\\
				\bottomrule
			\end{tabular*}%
		\end{threeparttable}
	}
	\label{tab:ssd}%
\end{table}%
\begin{figure}
	\centering
	\includegraphics[width=1.0\linewidth]{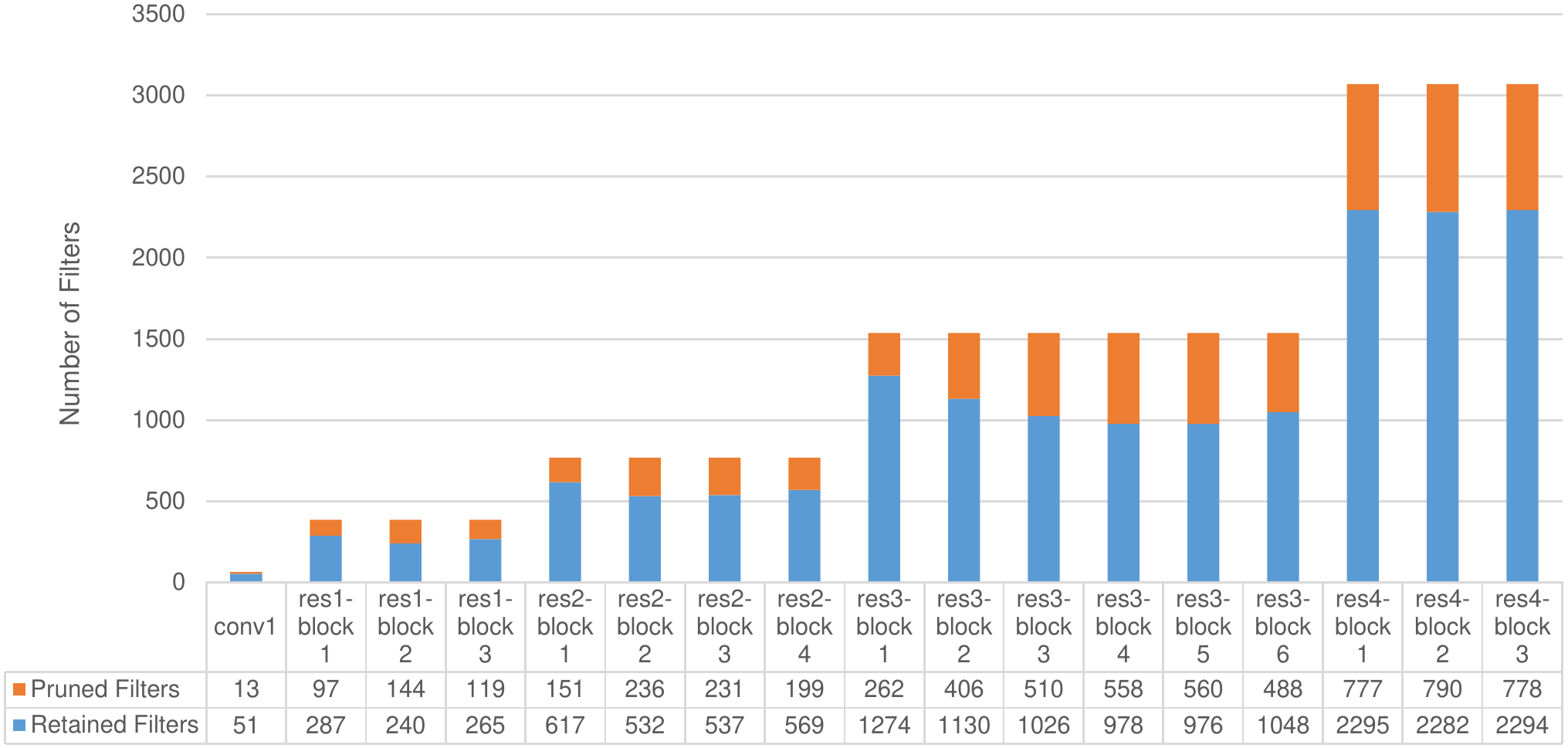}
	\caption{Statistics of pruned (orange) and retained (blue) filters of ResNet50 on ImageNet with Ours-0.2 method. Shortcuts with convolutional layers are not included.}
	\label{fig:filteranalysis}
\end{figure}

%-----------------------------------------------------------
\subsection{Results on Object Detection Task}
The proposed method is also applicable to other major computer vision tasks such as object detection. We take the SSD~\cite{ssd} (PyTorch version) as an example, its backbone network is similar to VGG16 and we prune it with VOC0712 dataset. Table.~\ref{tab:ssd} shows that our method is effective for object detection task, for example, it removes 30.92\% FLOPs and 34,80\% Parameters from baseline with imporvement of 0.1\% mAP. In fact, it is widely applicable to any network with common CNN structures.

%-----------------------------------------------------------
\subsection{Filter Pruning Analysis}
As shown in Fig.~\ref{fig:filteranalysis}, we reveal the pruning details about Ours-0.2 result on ImageNet dataset. We count the number of pruned/retained filters in all the layers except the shortcut connections with convolutional filters of ResNet50. For simplicity, the filter numbers of layers in the same block have been added together, the full result of each layer can be found in our supplementary material.

In ResNet50, there are 16 residual-blocks and one convolutional layer. Fig.~\ref{fig:filteranalysis} shows that the pruned filters mainly distribute in the high-level blocks. The filters of the high-level blocks contain more semantic information in detail, some of which are redundant. Empirically, removing these filters has less impact on the performance of the network. Meanwhile, the pruning rate of each block is different as illustrated in Fig.~\ref{fig:filteranalysis}, which also proves that our method attempts to search for the optimal pruning rate in each layer.

%===========================================================
\section{Conclusions}
In this paper, we propose a method to search the optimal pruning rate in a layer-wise manner and only needs fine-tuning for once. Based on the filter importance criterion derived from loss variation and first-order approximation, convolutional networks can be pruned efficiently. The group pruning and channel selection mechanism are also introduced to adapt with shortcut connections in networks. For practical usage, binary search the threshold accelerates pruning at a given global pruning rate for the entire network without extra fine-tuning. Experiments demonstrate that our method outperforms previous state-of-the-art pruning methods on different datasets and networks. The code is available at \url{https://github.com/Nuctech-AI/LBS_pruning}.

%===========================================================
\bibliographystyle{splncs}
\bibliography{egbib}

%this would normally be the end of your paper, but you may also have an appendix
%within the given limit of number of pages
\end{document}